\documentclass[runningheads]{llncs}

\usepackage[colorlinks,
            linkcolor=blue,
            anchorcolor=blue,
            citecolor=blue]{hyperref}

\usepackage{amsmath,bm}
\usepackage{amssymb}
\usepackage{amsfonts}
\usepackage{multirow}
\usepackage{booktabs}
\usepackage{threeparttable}
\usepackage{enumerate}
\usepackage[misc]{ifsym} 
\usepackage[pdftex]{graphicx}
\DeclareGraphicsExtensions{.pdf,.jpeg,.png}
\usepackage{color}

\begin{document}
\title{DSR: Direct Simultaneous Registration for Multiple 3D Images\thanks{This paper has been accepted by MICCAI 2022 Conference}}
%
%
\author{Zhehua Mao\inst{1} \and
Liang Zhao\inst{1}\textsuperscript{(\Letter)} \and
Shoudong Huang\inst{1} \and
Yiting Fan\inst{2}  \and \\
Alex P.W. Lee\inst{3}}
%
\authorrunning{Z. Mao et al.}

\institute{Robotics Institute, Faculty of Engineering and Information Technology, University of Technology Sydney, Ultimo, NSW 2007, Australia \\
\email {Liang.Zhao@uts.edu.au}\and
Department of Cardiology, Shanghai Chest Hospital, Shanghai Jiao Tong University, Shanghai, China \and
Division of Cardiology, Department of Medicine and Therapeutics, Prince of Wales Hospital and Laboratory of Cardiac Imaging and 3D Printing, Li Ka Shing Institute of Health Science, Faculty of Medicine, The Chinese University of Hong Kong, Hong~Kong, China}

\maketitle
%

\begin{abstract}

This paper presents a novel algorithm named Direct Simultaneous Registration (DSR) that registers a collection of 3D images in~a simultaneous fashion without specifying any reference image, feature extraction and matching, or information loss or reuse. The algorithm optimizes the global poses of local image frames by maximizing the similarity between a predefined panoramic image and local images. Although we formulate the problem as a Direct Bundle Adjustment (DBA) that jointly optimizes the poses of local frames and the intensities of the panoramic image, by investigating the independence of pose estimation from the panoramic image in the solving process, DSR is proposed to solve the poses only and proved to be able to obtain the same optimal poses as DBA. The proposed method is particularly suitable for the scenarios where distinct features are not available, such as Transesophageal Echocardiography (TEE) images. DSR is evaluated~by comparing it with four widely used methods via simulated and in-vivo 3D TEE images. It is shown that the proposed method outperforms these four methods in terms of accuracy and requires much fewer~computational resources than the state-of-the-art accumulated pairwise estimates (APE). Codes of DSR are available at \url{https://github.com/ZH-Mao/DSR}.
\end{abstract}

\section{Introduction}

Image registration is a fundamental task for many medical image analysis problems where valuable information conveyed by two or more images needs to be combined and examined \cite{oliveira2014medical,wachinger2012simultaneous}. In recent decades, mainstream medical imaging techniques, such as CT, MRI, and Ultrasound (US), have evolved from 2D to 3D, which proposes new challenges to medical image registration, such as feature extraction and high computational complexity \cite{schneider2012real}. Compared to feature-based methods \cite{ni2009reconstruction}, direct (intensity-based) methods \cite{oliveira2014medical,szeliski2006image} have occupied a dominant position in the field of medical image registration~\cite{hill2001medical,VIERGEVER2016140} because of their avoidance of feature extraction and high accuracy, especially when handling images that lack distinct features, such as 3D Transesophageal Echocardiography (3D TEE) images. Direct methods estimate the frame poses by maximizing the similarity between the images. Widely used similarity metrics include sum-of-squared differences (SSD), correlation ratio (CR), and mutual information (MI)~\cite{oliveira2014medical}.

Registration of a collection of images is much more complex than pairwise registration \cite{wachinger2012simultaneous}. One solution to this problem is to deduce the global poses from the results of pairwise registration \cite{szeliski2006image,wachinger2007three,carminati2015reconstruction,mao2021direct}. This strategy, although intuitive, is usually biased to the selected reference image and inevitably brings in accumulating errors. In comparison, a better strategy is to optimize the poses of all local frames simultaneously to avoid biases. In \cite{learned2005data}, a framework called congealing is proposed, which uses underlying entropic information of images for alignment. A large number of images are necessary for congealing because the estimation is done with the information at one location at a time \cite{wachinger2012simultaneous}. And as~\cite{cox2009least} pointed out, employing entropy for congealing is problematic due to its poor optimization characteristics. Recently in \cite{wachinger2012simultaneous}, an accumulated pairwise estimates (APE) method is proposed for simultaneous registration. The method considers overlapping areas of images in the objective function multiple times, thus may have the information reuse issue and bring in extra complexity in the optimization.

In this paper, we propose a novel direct simultaneous registration (DSR) method which optimizes global poses of a collection of 3D images directly based on image intensity. The novelties of the paper include: 1) simultaneous registration is formulated as a direct bundle adjustment (DBA) problem, which redefines classical bundle adjustment (BA) \cite{triggs1999bundle} by jointly optimizing the poses of local frames and the intensities of the predefined panorama; 2) DBA uses intensity information directly instead of the extracted and matched feature points of the local 2D images in classical BA. Therefore, our method can deal with images lacking distinct features such as 3D TEE images; 3) importantly, we prove in DBA, the pose estimation is independent of the intensities of the panorama during the optimization process; 4) based on 3), we derive DSR that \textit{only} solves the poses \textit{without} solving the intensities of the panorama but obtaining the same poses as DBA. Simulated and in-vivo 3D TEE images are used to evaluate the proposed DSR method compared with pairwise~\cite{carminati2015reconstruction}, Lie normalization \cite{wachinger2007three}, sequential \cite{mao2021direct}, and APE \cite{wachinger2012simultaneous} methods. Running in a simultaneous fashion, DSR is an unbiased method that can employ all intensity information of images without information reuse, which is an elegant way to obtain the optimal poses of local frames with high accuracy.

\section{Methodology}

\subsection{Direct Bundle Adjustment}

Suppose there are $m$ frames of 3D images taken from different viewpoints denoted as ${\bf I} = \{I_1, ..., I_i, ..., I_m\}$. Correspondingly, the rigid transformation for each frame is parameterized in Lie algebra space~\cite{hall2015lie} with the pose parameters ${\bf x}_\xi=[{\bm \xi}_1^{\top}, ...,{\bm \xi}_i^{\top}, ..., {\bm \xi}_m^{\top}]^\top \in \mathbb{R}^{6m}$. Simultaneous registration is the process of estimating the optimal pose parameters of all local frames $\hat{\bf x}_\xi$ simultaneously in order to align all the images in one global coordinate frame.

Assume $M$ is defined as a 3D panoramic image in the global frame. $M$ consists of $n$ voxels $\{{\bf p}_1, ..., {\bf p}_j, ..., {\bf p}_n\}$ where ${\bf p}_j\in \mathbb{R}^{3}$, which fuses all the local frame images. The intensity of voxel ${\bf p}_j$ in $M$ is obtained from fusing different points' intensities in local images. Denote the intensity of ${\bf p}_j$ in $M$ and ${\bf p}_j$'s corresponding point ${\bf p}_{ij}$ in local frame $I_i$ as $M({\bf p}_j)$ and  $I_i({\bf p}_{ij})$, respectively. The intensity difference between $M({\bf p}_{j})$ and $I_i({\bf p}_{ij})$ is
\begin{align}\label{Eq1}
  e_{ij}({\bm \xi}_{i}, M({\bf p}_{j})) &=M({\bf p}_{j})-I_i(\omega({\bm \xi_i},{\bf p}_{j}))= M({\bf p}_{j})-I_i({\bf p}_{ij}),
\end{align}
where ${\bf p}_{ij} =\omega({\bm \xi}_i,  {{\bf p}_j})={T}({\bm \xi}_i){\bf p}_{j}$ transforms ${\bf p}_j$ to ${\bf p}_{ij}$, and  ${ T}{(\cdot)} \in {{SE}(3)}$ maps the pose parameters ${\bm \xi}_i$ to a 3D Euclidean transformation. When calculating the intensity difference $e_{ij}$  in (\ref{Eq1}), the intensity of ${\bf p}_{ij}$ in $I_i$ is obtained using trilinear interpolation to reduce the error of the intensity difference computation.

Inspired by the conventional BA framework that considers both the 3D point positions and camera poses in the optimization~\cite{triggs1999bundle,alismail2016photometric}, we propose the DBA framework that jointly optimizes the poses of local frames and the intensities of the panoramic image (instead of 3D point positions in~BA). The overall state parameters considered in DBA are ${\bf x} = [{\bf x}^{\top}_{\xi}, {\bf x}^\top_{M}]^\top $, where $ {\bf x}_{M}=[ M({\bf p}_1), ... , M({\bf p}_j), ..., M({\bf p}_n)]^\top$  are the intensities of voxels in $M$. Then, we seek to obtain the optimal solution $\hat{\bf x} = [\hat{\bf x}^{\top}_{\xi}, \hat{\bf x}^\top_{M}]^\top $ that minimizes the sum-of-squared intensity differences between the panoramic image and the local images, i.e.
\begin{equation}\label{Eq2}
  {\bf \hat{x}} =\underset{ {{\bf x}_{\xi}, {\bf x}_M}}{\operatorname{argmin}} \sum_{j=1}^{n} \sum_{i=1}^{m}\sigma {({\bf p}_{ij})}(e_{ij}({\bm \xi}_{i}, M({\bf p}_{j})))^{2},
\end{equation}
where $\sigma( {{\bf p}_{ij}})=1$ if the transformed point {${\bf p}_{ij}$} is within Image $I_i$, i.e. ${\bf p}_j$ is observed in $I_i$, otherwise $\sigma( {{\bf p}_{ij}})=0$.
Such a formulation of DBA circumvents the process of feature extraction and matching in most BA problems. Additionally, since we take the intensities of the panoramic image into account, DBA can also obtain the optimal panoramic image besides the global poses of local frames.

Gauss-Newton (GN) method is commonly used to solve nonlinear least-squares (NLLS) problems like (\ref{Eq2}). The method obtains the solution by starting~with parameters initialization and then updating the parameters using the step changes calculated from GN equation in each iteration until the algorithm converges. If we write the overall observed intensity differences as a concatenation vector $e({\bf x}) = [..., e_{ij}, ...]^\top$, the objective function of (\ref{Eq2}) can be rewritten as
$f({\bf x}) = e({\bf x})^\top e({\bf x}).$
And step changes $\Delta {\bf x}$ in each iteration can be calculated from the GN equation:
\begin{equation}\label{Eq3}
J({\bf x})^\top J({\bf x}) \Delta {\bf x} = -J({\bf x})^\top e({\bf x}),
\end{equation}
where $J({\bf x})$ is the Jacobian matrix of $e({\bf x})$ w.r.t. $\bf x$.

Let $J_{ij}({\bf x})$ denote one row of $J(\bf x)$, which is the gradient of one intensity difference $e_{ij}$ w.r.t. ${\bf x} = [{\bf x}^{\top}_{\xi}, {\bf x}^\top_{M}]^\top $.
It is shown in (\ref{Eq1}) that $e_{ij}$ is only dependent on ${\bm \xi}_i$ and $M({\bf p}_j)$, thus only two blocks in $J_{ij}({\bf x})$ are nonzero, i.e.
\begin{equation*}
\frac{\partial e_{ij}({{\bm \xi}}_{i}, M({\bf p}_{j}))}{\partial{\bm \xi}_{i}} =-\frac{\partial I_i}{\partial \omega({\bm \xi}_i,{\bf p}_j)} \frac{\partial \omega({\bm \xi}_i,{\bf p}_j)}{\partial {\bm \xi}_{i}},~\text{and}~
\frac{\partial e_{ij}({{\bm \xi}}_{i}, M({\bf p}_{j}))}{\partial (M({\bf p}_{j}))} = 1,
\end{equation*}
which indicates $J_{ij}({\bf x})$ is very sparse.

Although the optimal poses and panoramic image can be obtained simultaneously, DBA seems more difficult to solve than traditional multi-image registration problems since a much higher order state vector is involved. However, we can further prove that the pose optimization is actually independent of the panoramic image in the GN iterations (see Section \ref{DSR}), which means we do not need to solve the intensities of the panoramic image but can obtain exactly the same optimal poses as solving the complete DBA. This is also an important property that conventional BA frameworks do not have.

\subsection{Simultaneous Registration without Intensity Optimization}\label{DSR}

\noindent{\it {\bf Theorem}: {When solving (\ref{Eq2}) with GN iterations, the optimization of poses is independent of the intensities of the panoramic image.}}

\noindent{\bf Proof}:
If we write Jacobian matrix of $e({\bf x})$ w.r.t. $\bf x_{ \xi}$ and ${\bf x}_{M}$ separately as $J({\bf x})= [J_{\xi}, J_M]$, then (\ref{Eq3}) can be rewritten as the following format:

\begin{equation}\label{Eq4}
\left[\begin{array}{cc}
H_{{ \xi}{ \xi}} & H_{{ \xi}M} \\
H_{{M} \xi} & H_{{M}M}
\end{array}\right]\left[\begin{array}{l}
\Delta {\bf x}_{ \xi} \\
\Delta {\bf x}_{M}
\end{array}\right]=\left[\begin{array}{l}
b_{\xi} \\
b_M
\end{array}\right],
\end{equation}
where $H_{{ \xi}{ \xi}}=J_{ \xi}^{\top}J_{ \xi}, H_{{ \xi}M}= H^{\top}_{{M} \xi}  =J_{ \xi}^{\top}J_M, H_{{M}M}=J_{M}^{\top}J_M, b_{\xi}=-J_{ \xi}^{\top}e({\bf x})$, and $ b_M= -J_{M}^{\top}e({\bf x})$.
Then, through Schur complement \cite{zhang2006schur}, the step changes of poses and intensities of voxels can be computed sequentially as:
\begin{align}
  \left(H_{{ \xi}{ \xi}}-H_{{ \xi}M}H_{{M}M}^{-1} H_{{ \xi}M}^{\top}\right)\Delta {\bf x}_{ \xi}&=(b_{ \xi}-H_{{ \xi}M}H_{{M}M}^{-1} b_M), \label{Eq5}\\
  H_{{M}M}\Delta {\bf x}_{M}&=b_M-H_{{ \xi}M}^{\top} \Delta {\bf x}_{ \xi}. \label{Eq6}
\end{align}

Suppose intensity differences $e({\bf x})$ are decomposed into two components $e({\bf x}) = {\bf A} -{\bf B}$, where ${\bf A} = [..., M({\bf p}_j), ...]^\top$ and ${\bf B}=[..., I_i({\bf p}_{ij}), ...]^\top$ represent observed intensities of the panoramic image and their corresponding intensities in local frames, respectively. The right side of (\ref{Eq5}) becomes:
\begin{align}\label{Eq7}
 &b_{ \xi}-H_{{ \xi}M}H_{{M}M}^{-1} b_M
 =-J_{{ \xi}}^{\top}({\bf A}-{\bf B})+J_{ \xi}^{\top}J_{M}(J_{M}^{\top}J_{M})^{-1} J_{M}^{\top}({\bf A}-{\bf B}) \nonumber\\
 &=-J_{{ \xi}}^{\top}({\bf A}-J_{M}(J_{M}^{\top}J_{M})^{-1} J_{M}^{\top}{\bf A})-J_{ \xi}^{\top}(J_{M}(J_{M}^{\top}J_{M})^{-1} J_{M}^{\top}{\bf B}-{\bf B}).
\end{align}
It is shown from $J_{ij}({\bf x})$ that there is one and only one nonzero element 1 in each row of $J_M$. The nonzero element means the voxel ${\bf p}_j$ is observed in the local frame $i$ and corresponds to the intensity difference $e_{ij}= M({\bf p}_{j})-I_i({\bf p}_{ij})$. Therefore, according to the observed status of the panoramic image in the local frames which is indicated by the structure of $J_{M}$, it can be easily deduced that $ {\bf A}=J_M{\bf x}_M $. Substituting $ {\bf A}$ to the first term on the right side of (\ref{Eq7}), we have:
\begin{align}\label{Eq8}
-J_{ \xi}^{\top}(J_M{\bf x}_M-J_{M}(J_{M}^{\top}J_{M})^{-1} J_{M}^{\top}{J_M{\bf x}_M}) = {\bf 0}.
\end{align}
Then, (\ref{Eq5}) becomes:
\begin{align}\label{Eq9}
\left(H_{{ \xi} \xi}-H_{{ \xi}M}H_{{M}M}^{-1} H_{{ \xi}M}^{\top}\right)\Delta {\bf x}_{ \xi}= -J_{ \xi}^{\top}(J_{M}H_{{M}M}^{-1}J_{M}^{\top}{\bf B}-{\bf B}),
\end{align}
{which indicates that the step change $\Delta {\bf x}_{ \xi}$ is independent of intensities ${\bf x}_M$ in every GN iteration. Therefore, obtaining the optimal poses is independent of the intensities of panoramic image $M$ during the optimization process.}
\hfill {\bf Q.E.D.}

Although $H_{MM}$ in (\ref{Eq9}) has a huge dimension due to the large number of intensities in $M$, $H_{MM}$ is sparse and diagonal because of the sparse structure of $J_M$. {The value of a diagonal element represents the number of times that the corresponding voxel of $M$ has been observed in the local frames in the current iteration.} So, the inverse of diagonal matrix $H_{MM}$  can be easily computed by {finding the inverse of each diagonal element}, which makes solving (\ref{Eq9}) efficient.

Such an independent property of DBA is very attractive since it allows us to optimize the poses only using (\ref{Eq9}), which is equivalent to solving the complete DBA problem using (\ref{Eq4}). A 3D image typically contains millions of voxels but we only need six parameters to represent its pose.
Therefore, the independence of optimizing poses to intensities can greatly help us reduce the dimension of the solution space. For distinction with DBA, we call this method DSR.

In addition, after $\hat{\bf x}_{\xi}$ is obtained, the NLLS problem in (\ref{Eq2}) becomes a linear least-squares problem. Therefore, if required, we can calculate the optimal panoramic image easily in only one step from the following closed-form formula:
\begin{equation}\label{opt_intensity}
\hat{\bf x}_{M} = - H_{{M}M}^{-1}J_{M}^{\top}{\bf B}.
\end{equation}
The implementation process of DSR is summarized in Algorithm 1 in the supplementary materials.

\section{Experiments and Results}
Registration of US images is usually more challenging than other modalities like CT and MRI due to the relatively low signal-to-noise ratio. Additionally, registration of a collection of 3D TEE images is especially valuable to overcome the drawback of small field of view (FoV) of 3D TEE probes. Thus, in this section, 3D TEE images are used as examples to evaluate the proposed DSR algorithm compared with the pairwise \cite{carminati2015reconstruction}, Lie normalization \cite{wachinger2007three}, sequential \cite{mao2021direct}, and APE~\cite{wachinger2012simultaneous} methods. {Both simulated and in-vivo experiments are performed}.

\subsection{Simulated Experiments}

Five sequences of 3D images (in grayscale ranging from 0 to 255) are simulated by transforming a 3D TEE volume to a real 3D CT scan of the heart with different poses and cropping the corresponding image area. The FoV of the simulated TEE images is the same as the 3D TEE image to get similar imagery information as real TEE volumes. Each sequence contains 11 frames of 3D images. The magnitude of these transformations varies between +/-- 12 degrees for rotations and +/-- 15 pixels for translations, which are typical ranges of poses in our obtained in-vivo 3D TEE images. Multiplicative speckle noise \cite{loizou2005comparative} in the original US signal can be transformed into a kind of additive noise close to Gaussian distribution in the obtained US images after logarithmic compression \cite{loizou2005comparative,zhang2020despeckling}. Thus, Gaussian noise with a standard deviation of 25 is generated randomly and added to the intensities of these five~sequences of images.

\begin{figure}[t]
	\centering
	\includegraphics[width=4.6in]{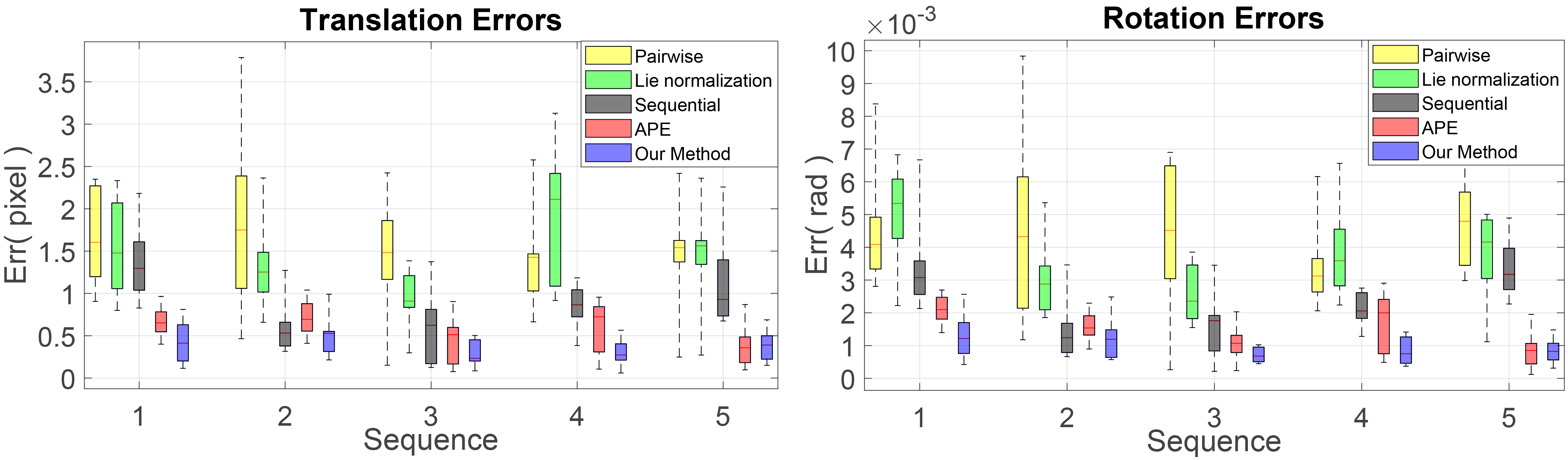}
	\caption{{Accuracy of DSR method compared to pairwise \cite{carminati2015reconstruction}, Lie normalization \cite{wachinger2007three}, sequential \cite{mao2021direct}, and APE \cite{wachinger2012simultaneous} methods using 5 sequences of simulated 3D TEE images.}} \label{Fig1}
\end{figure}

The accuracy of the proposed DSR method is evaluated by comparing it with pairwise, Lie normalization, sequential, and APE methods via simulated datasets. For fair comparisons, pairwise, sequential, and APE methods use the SSD as the similarity metric, GN method as optimization solver, and the initial pose parameters in each method are the same. Lie normalization optimizes poses obtained from pairwise methods and does not directly involve images~\cite{wachinger2007three}. Thus, we use the results from the pairwise registration as the input to Lie normalization. The mean absolute errors (MAE) of translation and Euler angles obtained from the proposed DSR and other four methods are compared in Fig.~\ref{Fig1}.

It is shown from Fig. \ref{Fig1} that MAE of the results obtained from DSR, sequential, and APE methods are much smaller than the pairwise and Lie normalization methods in most of the cases, which indicates the better accuracy of these three methods. Additionally, among DSR, sequential, and APE methods, the accuracy of DSR is within 0.5 pixels for translations in most of the cases and is within $1 \times 10^{-3}$ rad for rotations in more than half cases. Although the accuracy of APE is the closest to DSR among four competing methods, it still has larger errors than DSR. The errors of APE are greater than 0.5 pixels for translations and greater than $1 \times 10^{-3}$ rad for rotations in more than half of experiments. And the results show that the accuracy of both translations and rotations from the sequential method is lower than DSR and APE. Furthermore, it is seen from Fig.~\ref{Fig1} that the distribution of errors from DSR is more concentrated than the others, which indicates it also has better robustness than the other four methods.

The above comparisons indicate that the proposed method has the highest accuracy, followed by APE. Both these two simultaneous registration methods are more accurate than the other three which are deduced from pairwise registration. In addition, compared with the proposed DSR method, one apparent drawback of APE is its much higher computational complexity. Since both DSR and APE use the sum-of-squared intensity differences as the objective function, the computational complexities for both methods are closely related to the number of intensity differences. Suppose there are $m$ images, each image has $h$ pixels, and every two images have around $\alpha$\% overlapping area, the computational complexity of APE is around {\textbf{\textit O}}$(m(m-1)/2\times \alpha\% \times h)$ \cite{wachinger2010structural} since APE considers all the combinations of images, while that of DSR is only {\textbf{\textit O}}$(m\times h)$. In the simulated experiments for each sequence, it is found that APE calculates around four times as many intensity differences as DSR and needs around 4-5 times longer time than DSR for each iteration. Theoretically, the more images involved, the higher the computational complexity of APE is, and the more time it takes than~DSR.

\subsection{In-vivo Experiments}

In the in-vivo experiments, forty-six 3D TEE images from six patients (Patient $\#1$ to $\#6$) are collected using an iE33 ultrasound system (Philips Medical Systems) equipped with an X7-2 real-time 3D transducer. ECG-gating technique~\cite{fenster2001three} is used to assist capture ECG-gated 3D TEE images so that registration of these images can be considered as rigid. The number of images in each dataset varies from 6 to 11. Each 3D TEE image contains around six million voxels. Since the proposed DSR, sequential, and APE methods outperformed the other two methods in terms of accuracy, in this section, the proposed method is compared with the sequential and APE methods only. Pose parameters of three methods are initialized using the results from the pairwise method. It is found that such an initialization method is enough for DSR to converge to the correct results.

Analyzing the accuracy of a registration algorithm based on in-vivo datasets is complex because the ground-truth poses are usually not available. If images are aligned using the estimated poses, visually we can confirm that the stitching areas of the aligned images should be smooth and without misalignment if the poses are accurate. Therefore, to evaluate the accuracy of the proposed DSR method, pairwise images are aligned using the poses obtained by the sequential, APE, and DSR methods, respectively.

\begin{figure*}[t]
	\centering
	\includegraphics[width=4.5in]{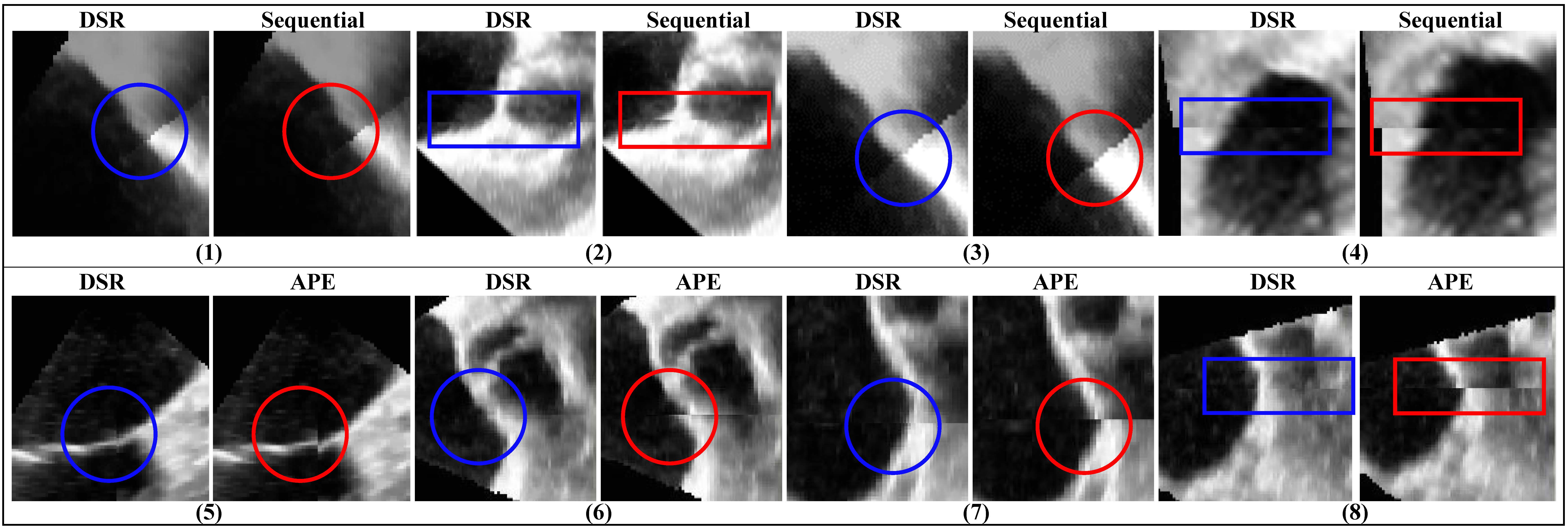}
	\caption{Comparisons of the aligned images using poses from sequential, APE, and DSR.} \label{Fig2}
\end{figure*}

Similar to the results in our simulated experiments, aligned images obtained from DSR have the best quality, followed by APE and then the sequential method. In all the aligned images, there is no misalignment found from the proposed DSR in the experiments, while some apparent misalignment is found from the results of APE and sequential methods. Several examples are displayed in Fig.~\ref{Fig2}. Additionally, although APE can obtain results that are closer to those from the proposed method than the sequential method, it requires a much longer time for each iteration. In the experiments, APE usually needs around 2-5 times longer time than the proposed method for each iteration.

\begin{figure*}[t]
	\centering
	\includegraphics[width=4.7in]{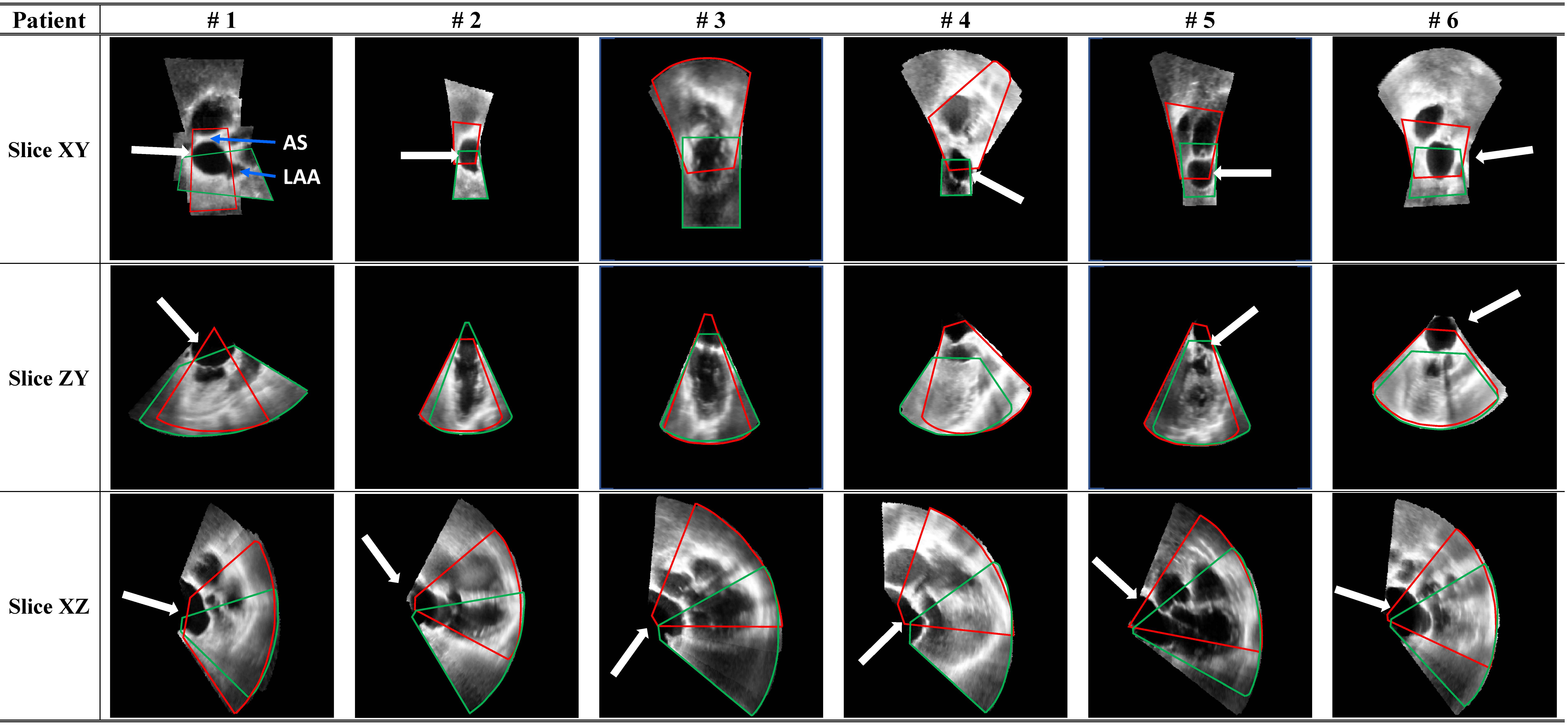}
	\caption{Fused 3D TEE images using registration results from DSR for six in-vivo datasets. LA walls which have sharp structures in the images are indicated by white arrows in selected areas. Colored frames are the boundaries of two registered volumes.} \label{Fig3}
\end{figure*}

To further evaluate the accuracy of the proposed method, in-vivo 3D~TEE images in each dataset are fused using (\ref{opt_intensity}) and the estimated poses from DSR. We manually select areas which contain sharp boundaries like the left atrium (LA) wall in the fused images for evaluation since generally, misalignment caused by poses with low accuracy can be easily found in these areas. The selected regions are shown in Fig. \ref{Fig3} with three orthogonal slices and two of the registered images in the fused images are highlighted in color boundaries. By observing the LA walls which are indicated by white arrows in Fig. \ref{Fig3}, it is shown that the stitching areas have smooth transition and no misalignment is found in the images, which suggests good quality of alignments have been obtained by DSR.

The motivation for our current study is to enlarge the FoV of 3D TEE to assist transcatheter left atrial appendage (LAA) occlusion \cite{fan2019device}. First, the enlarged FoV of 3D TEE allows the LAA to be observed completely (see Patient \# 1 in Fig. \ref{Fig3}) to facilitate device size selection for LAA occlusion \cite{morais2019semiautomatic}. Secondly, a complete structure of the left atrium in the enlarged 3D TEE image allows measuring the relative position and orientation of LAA w.r.t. atrial septum (AS) to facilitate the planning for LAA occlusion \cite{fan2019device}. By counting the number of voxels, it is found that the FoV of the fused image is enlarged to 2.18, 2.10, 2.02, 2.01, 1.80, and 2.06 times as compared with the original single TEE volume of Patient \#~1 to \# 6, respectively.

\section{Conclusion}
Starting from the framework of direct bundle adjustment, a novel direct simultaneous registration algorithm for 3D images is proposed in the paper. The method can optimize the poses of a collection of local images simultaneously without any information loss or reuse. Results from the simulated and in-vivo experiments demonstrate that the proposed method outperforms the other four competing methods in terms of accuracy and is more efficient than the state-of-the-art APE method. From the results of simulated experiments, it is evident that our method improved the accuracy of registration by more than 50\% compared to the other four methods for most cases. In-vivo experiments also show accurate structures and extended field of view of the fused images, indicating a good quality of registration and a significant potential clinical value of the proposed method.

The proposed method can be very useful in practice when real-time performance is not required, e.g. using the enlarged FoV of 3D TEE for surgical planning of LAA occlusion. Our current focus is more on accuracy than efficiency, thus we implemented DSR in MATLAB on CPU. Since the linear system~(\ref{Eq4}) has a special sparse structure that is similar to other bundle adjustment problems, it is very promising for us to use techniques in \cite{wu2011multicore} to achieve parallel implementation of DSR on GPU. Additionally, optimization techniques used in g2o~\cite{kummerle2011g} and parallax BA~\cite{zhao2015parallaxba} could also help us achieve the fast implementation. Our future work will focus on the efficient implementation of DSR.

\bibliographystyle{splncs04}
\bibliography{./references.bib}

\begin{thebibliography}{10}
\providecommand{\url}[1]{\texttt{#1}}
\providecommand{\urlprefix}{URL }
\providecommand{\doi}[1]{https://doi.org/#1}

\bibitem{alismail2016photometric}
Alismail, H., Browning, B., Lucey, S.: Photometric bundle adjustment for
  vision-based slam. In: Asian Conference on Computer Vision. pp. 324--341.
  Springer (2016)

\bibitem{carminati2015reconstruction}
Carminati, M.C., Piazzese, C., Weinert, L., Tsang, W., Tamborini, G., Pepi, M.,
  Lang, R.M., Caiani, E.G.: Reconstruction of the descending thoracic aorta by
  multiview compounding of 3-d transesophageal echocardiographic aortic data
  sets for improved examination and quantification of atheroma burden.
  Ultrasound in Medicine \& Biology  \textbf{41}(5),  1263--1276 (2015)

\bibitem{cox2009least}
Cox, M., Sridharan, S., Lucey, S., Cohn, J.: Least-squares congealing for large
  numbers of images. In: 2009 IEEE 12th International Conference on Computer
  Vision. pp. 1949--1956. IEEE (2009)

\bibitem{fan2019device}
Fan, Y., Yang, F., Cheung, G.S.H., Chan, A.K.Y., Wang, D.D., Lam, Y.Y., Chow,
  M.C.K., Leong, M.C.W., Kam, K.K.H., So, K.C.Y., et~al.: Device sizing guided
  by echocardiography-based three-dimensional printing is associated with
  superior outcome after percutaneous left atrial appendage occlusion. Journal
  of the American Society of Echocardiography  \textbf{32}(6),  708--719 (2019)

\bibitem{fenster2001three}
Fenster, A., Downey, D.B., Cardinal, H.N.: Three-dimensional ultrasound
  imaging. Physics in medicine \& biology  \textbf{46}(5), ~R67 (2001)

\bibitem{hall2015lie}
Hall, B.: Lie groups, Lie algebras, and representations: an elementary
  introduction, vol.~222. Springer (2015)

\bibitem{hill2001medical}
Hill, D.L., Batchelor, P.G., Holden, M., Hawkes, D.J.: Medical image
  registration. Physics in medicine \& biology  \textbf{46}(3), ~R1 (2001)

\bibitem{kummerle2011g}
K{\"u}mmerle, R., Grisetti, G., Strasdat, H., Konolige, K., Burgard, W.: g 2 o:
  A general framework for graph optimization. In: 2011 IEEE International
  Conference on Robotics and Automation. pp. 3607--3613. IEEE (2011)

\bibitem{learned2005data}
Learned-Miller, E.G.: Data driven image models through continuous joint
  alignment. IEEE Transactions on Pattern Analysis and Machine Intelligence
  \textbf{28}(2),  236--250 (2005)

\bibitem{loizou2005comparative}
Loizou, C.P., Pattichis, C.S., Christodoulou, C.I., Istepanian, R.S.,
  Pantziaris, M., Nicolaides, A.: Comparative evaluation of despeckle filtering
  in ultrasound imaging of the carotid artery. IEEE transactions on
  ultrasonics, ferroelectrics, and frequency control  \textbf{52}(10),
  1653--1669 (2005)

\bibitem{mao2021direct}
Mao, Z., Zhao, L., Huang, S., Fan, Y., Lee, A.P.W.: Direct 3d ultrasound fusion
  for transesophageal echocardiography. Computers in Biology and Medicine
  \textbf{134},  104502 (2021)

\bibitem{morais2019semiautomatic}
Morais, P., Vila{\c{c}}a, J.L., Queir{\'o}s, S., De~Meester, P., Budts, W.,
  Tavares, J.M.R., D’hooge, J.: Semiautomatic estimation of device size for
  left atrial appendage occlusion in 3-d tee images. IEEE transactions on
  ultrasonics, ferroelectrics, and frequency control  \textbf{66}(5),  922--929
  (2019)

\bibitem{ni2009reconstruction}
Ni, D., Chui, Y.P., Qu, Y., Yang, X., Qin, J., Wong, T.T., Ho, S.S., Heng,
  P.A.: Reconstruction of volumetric ultrasound panorama based on improved 3d
  sift. Computerized medical imaging and graphics  \textbf{33}(7),  559--566
  (2009)

\bibitem{oliveira2014medical}
Oliveira, F.P., Tavares, J.M.R.: Medical image registration: a review. Computer
  methods in biomechanics and biomedical engineering  \textbf{17}(2),  73--93
  (2014)

\bibitem{schneider2012real}
Schneider, R.J., Perrin, D.P., Vasilyev, N.V., Marx, G.R., Pedro, J., Howe,
  R.D.: Real-time image-based rigid registration of three-dimensional
  ultrasound. Medical image analysis  \textbf{16}(2),  402--414 (2012)

\bibitem{szeliski2006image}
Szeliski, R.: Image alignment and stitching: A tutorial. Foundations and
  Trends{\textregistered} in Computer Graphics and Vision  \textbf{2}(1),
  1--104 (2006)

\bibitem{triggs1999bundle}
Triggs, B., McLauchlan, P.F., Hartley, R.I., Fitzgibbon, A.W.: Bundle
  adjustment—a modern synthesis. In: International workshop on vision
  algorithms. pp. 298--372. Springer (1999)

\bibitem{VIERGEVER2016140}
Viergever, M.A., Maintz, J.A., Klein, S., Murphy, K., Staring, M., Pluim, J.P.:
  A survey of medical image registration – under review. Medical Image
  Analysis  \textbf{33},  140--144 (2016), 20th anniversary of the Medical
  Image Analysis journal (MedIA)

\bibitem{wachinger2010structural}
Wachinger, C., Navab, N.: Structural image representation for image
  registration. In: 2010 IEEE Computer Society Conference on Computer Vision
  and Pattern Recognition-Workshops. pp. 23--30. IEEE (2010)

\bibitem{wachinger2012simultaneous}
Wachinger, C., Navab, N.: Simultaneous registration of multiple images:
  similarity metrics and efficient optimization. IEEE transactions on pattern
  analysis and machine intelligence  \textbf{35}(5),  1221--1233 (2012)

\bibitem{wachinger2007three}
Wachinger, C., Wein, W., Navab, N.: Three-dimensional ultrasound mosaicing. In:
  International Conference on Medical Image Computing and Computer-Assisted
  Intervention. pp. 327--335. Springer (2007)

\bibitem{wu2011multicore}
Wu, C., Agarwal, S., Curless, B., Seitz, S.M.: Multicore bundle adjustment. In:
  CVPR 2011. pp. 3057--3064. IEEE (2011)

\bibitem{zhang2006schur}
Zhang, F.: The Schur complement and its applications, vol.~4. Springer Science
  \& Business Media (2006)

\bibitem{zhang2020despeckling}
Zhang, J., Cheng, Y.: Despeckling Methods for Medical Ultrasound Images.
  Springer (2020)

\bibitem{zhao2015parallaxba}
Zhao, L., Huang, S., Sun, Y., Yan, L., Dissanayake, G.: Parallaxba: bundle
  adjustment using parallax angle feature parametrization. The International
  Journal of Robotics Research  \textbf{34}(4-5),  493--516 (2015)

\end{thebibliography}

\end{document}